\documentclass[sigconf, nonacm]{acmart}
\usepackage{algorithmic}
\usepackage{balance}
\usepackage{algorithm}

\begin{document}

\title{Statistical Proof of Execution (SPEX)}

\newcommand{\sym}[1] {\textit{\textbf{#1}}}

\newtheorem{texample}{Example}[section]
\newtheorem{problem}{Problem}[section]

\author{Michele Dallachiesa}
\affiliation{\institution{Warden Labs}}

\author{Antonio Pitasi}
\affiliation{\institution{Warden Labs}}

\author{David Pinger}
\affiliation{\institution{Warden Labs}}

\author{Josh Goodbody}
\affiliation{\institution{Warden Labs}}

\author{Luis Vaello}
\affiliation{\institution{Warden Labs}}

\begin{abstract}
  Many real-world applications are increasingly incorporating automated decision-making, driven by the widespread adoption of ML/AI inference for planning and guidance. This study examines the growing need for verifiable computing in autonomous decision-making. We formalize the problem of verifiable computing and introduce a sampling-based protocol that is significantly faster, more cost-effective, and simpler than existing methods. Furthermore, we tackle the challenges posed by non-determinism, proposing a set of strategies to effectively manage common scenarios.
\end{abstract}

\maketitle

{\let\thefootnote\relax\footnote{{Correspondence to <michele@wardenprotocol.org>}}}

\section{Introduction}

As automated decision-making, frequently driven by ML/AI inference, becomes more embedded in critical systems, the demand for verifiable computations has grown significantly.
Lack of verification increases the risk of errors, biases, and manipulation.
This growing concern has driven research into enhancing the verifiability of computational processes to ensure that system outputs can be trusted.

We introduce the Statistical Proof of Execution (SPEX) protocol, a novel framework designed to provide strong and flexible compute correctness guarantees, including for workloads that may exhibit non-deterministic outputs, such as AI and LLMs, while offering comparatively lower overhead than other methods. SPEX ensures that computations are trustworthy and auditable, making it a crucial tool for developing secure and transparent systems.
Its ability to establish trust in computational outcomes is especially critical for applications in both cloud and blockchain environments, where transparency and accountability are paramount. Whether deployed in traditional industries or decentralized applications, verifiability acts as a proxy for ensuring quality and reliability.
In summary, the contributions of this paper are as follows:

\begin{itemize}
  \item We formally define the problem of verifiable computing.
  \item We propose a sampling-based protocol for verifiable computing that provides flexible guarantees with minimal computational and memory overhead, and full parallelizability. This approach is significantly faster, more cost-effective, and simpler than existing methods.
  \item We introduce extensions to verify non-deterministic computational states, such as floating-point arrays, semantically similar embeddings, and agentic plans, which are increasingly prevalent in autonomous decision-making.
  \item We provide an open-source reference implementation of our contributions in the \textit{warden-spex} Python package \cite{spex202501}.
\end{itemize}

This paper is organized as follows: Section~\ref{sec:relwork} reviews the related work, Section~\ref{sec:problemdef} formally defines the problem, while  Section~\ref{sec:drivingex} introduces a driving example. Section~\ref{sec:proposal} presents our solution, and Section~\ref{sec:conclusion} concludes the study.

\section{Related work}
\label{sec:relwork}

The challenges of verifiable, privacy-preserving, and trusted computing have been extensively studied in recent years. At their core, these problems arise from the need to delegate computation to third parties while ensuring correctness guarantees about its outputs.

Fully Homomorphic Encryption (FHE), Zero-Knowledge Proofs (ZKP), and Multi-Party Computation (MPC) are three cryptographic techniques that enable privacy-preserving computations but serve distinct purposes and operate under different paradigms. FHE allows computations on encrypted data, ZKP minimizes disclosure while proving correctness, and MPC leverages distributed computation to ensure privacy among participants.
\cite{xing2023zero} surveys the use of ZKPs for machine learning verifiability, while \cite{chen2024zkml} presents ZKML, an optimized system for performing machine learning inference within ZKP frameworks. \cite{knott2021crypten} introduces a framework with MPC primitives for machine learning functions, and \cite{pang2024bolt} focuses on optimizing MPC for transformer models. \cite{gilad2016cryptonets} explores the application of neural networks with encrypted data, achieving notable throughput and accuracy.
Despite their potential, FHE remains slow and resource-intensive, ZKP requires complex setup and produces large proofs, and MPC encounters communication challenges and scalability issues as the number of participants grows.

Trusted Execution Environments (TEEs) are secure areas within processors that ensure the confidentiality and integrity of sensitive data during processing, providing strong isolation from the main operating system. \cite{mo2024machine} provides a review of the existing TEE systems for ML use cases.
TEEs, despite their intended isolation features, may contain design or implementation vulnerabilities that can be exploited, as evidenced by past incidents, including 4 CVEs affecting Intel® SGX in 2024 alone.

Incentive-driven models employ random sampling, partial and complete re-execution to ensure correctness. These techniques are particularly effective against lazy actors, as any attempt to falsify a proof would require an effort at least equal to, or greater than, that of computing the correct answer, thereby negating any potential advantage. \cite{tan2017efficient} discusses the simultaneous replay and efficient verification of concurrent executions using web logs, while \cite{jia2021proof} introduces the concept of proof-of-learning in machine learning, which allows the verifier to recompute individual training steps to validate their correctness. \cite{arun2025verde} adapts refereed delegation for secure ML computation, ensuring correctness through arbitration and reproducibility mechanisms. The principles of proof of learning and proof of compute are further examined within the context of blockchain in \cite{atoma2024, zhang2024proof, zhao2024proof}, with \cite{conway2024opml} introducing an optimistic system that includes a challenger period.

Our proposal falls within the category of incentive-driven models, optimizing cost efficiency, overhead time, and statistical correctness guarantees while effectively handling non-determinism.

\section{Problem definition}
\label{sec:problemdef}
We will formalize the problem after introducing some definitions.
Solvers perform tasks that may subsequently be verified by verifiers:

\begin{itemize}
  \item \sym{Task} $t$: Any compute pipeline, including data transformations such as ETLs, ELTs, machine learning (ML) training workloads, AI inference, GenAI and LLM workloads.
  \item \sym{Specification} $f$: A function that is both a reference implementation and the correctness criterion for the outputs of task $t$ implementations.
  \item \sym{Solver} $S$: An executor for task $t$ that may have an incentive to manipulate and alter its outputs.
  \item \sym{Verifier} $V$: An executor that verifies the correctness of the outputs of solver $S$ according to specification $f$.
\end{itemize}

Execution and verification behaviors may vary:

\begin{itemize}
  \item \sym{Honest} execution ensures computations are carried out fully and correctly, maintaining accuracy and integrity.
  \item \sym{Lazy} execution skips or approximates steps to reduce effort, without strategic intent to alter results.
  \item \sym{Adversarial} execution manipulates computations to gain a strategic advantage, potentially incurring higher costs than honest computation by falsifying results, introducing bias, or selectively altering outputs.
  \item \sym{Dishonest} execution combines lazy and adversarial behaviors, aiming to reduce effort or gain strategic advantage.
\end{itemize}

We are now ready to formulate the verifiable computing problem:

\begin{problem}[Verifiable computing]
\label{problem:1}
Given task $t$ that is executed repeatedly with potentially varying inputs, function $s$ produces the output $r$ of task $t$ along with cryptographic proof $p$, which function $v$ can use to verify the correctness of $r$. Function $s$ runs on solver $S$, while $v$ runs on verifier $V$. Verifier $V$ is honest, whereas solver $S$ may be lazy or adversarial. The objective is to guarantee the correctness of $r$ with a confidence level of at least $\delta$, while minimizing the combined memory and computational costs of $s$ and $v$.
\end{problem}

In this study, we present a novel sampling-based approach that ensures robust and flexible statistical guarantees on solver output correctness while significantly reducing memory and computational costs compared to existing methods.

\section{Driving example}
\label{sec:drivingex}

To concretely illustrate the key concepts and methodologies discussed in this work, we present a driving example. This example serves as a representative case study, demonstrating the application of the theoretical framework in a practical context. It is designed to illustrate the protocol and does not cover non-determinism, which is discussed separately in Sections \ref{sec:hashing_fp}, \ref{sec:semhash} and \ref{sec:hashagent}.

\begin{texample}[Task \sym{PrimeSum}]
  The specification function $f$ takes as input an array of positive integers $i \geq 1$ and returns the sum of the prime numbers at the specified indices in the sequence of prime numbers. For example, given the input set $ \{ n \mid 1 \leq n \leq 10, \ n \in \mathbb{Z} \} $, the function retrieves the first 10 prime numbers: $[2, 3, 5, 7, 11, 13,$ $17, 19, 23, 29]$ and returns their sum, which is $129$.

\end{texample}

The implementation of task \sym{PrimeSum} is inherently parallelizable since the computation of individual primes for specific indices can be performed independently. These computed primes can then be aggregated efficiently using a map-reduce approach, where each parallel computation maps to a subset of primes, and their partial sums are reduced to produce the final result. A \sym{lazy} solver $S$ may approximate certain subsets of primes to reduce computational effort, whereas an \sym{adversarial} solver might compute the correct output but return a different one to gain a strategic advantage. The verifier $V$ is responsible for ensuring the correctness of the output with a confidence level of at least $\delta$.

\section{Proposed approach}
\label{sec:proposal}

In this section, we propose an efficient approach for evaluating verifiable computing.
Section~\ref{sec:baseline} presents the baseline protocol, while Section~\ref{sec:spex} introduces the \sym{SPEX} protocol.
Section~\ref{sec:samcomp} provides an overview of sampling strategies for computational states.
Section~\ref{sec:bloom} discusses the use of Bloom filters as cryptographic proofs of execution, covering techniques for hashing of numerical arrays, semantic hashing of embeddings, and hashing of agentic plans.

\subsection{The Baseline Protocol}
\label{sec:baseline}

We present the baseline protocol for verifiable computing, as introduced in Problem~\ref{problem:1}:

\vspace*{.1cm}
{\em The solver function $s$ computes the task $t$ using the reference implementation $f$ and outputs the result $r$ along with a proof $p$, where $p$ is equal to the result $r$. The verifier function $v$ independently recomputes $f$ to check whether the output matches, effectively replicating the computation as an honest solver and ensuring correctness with the highest confidence level of $\delta = 1$.}
\vspace*{.1cm}

When applying the baseline protocol to the \sym{PrimeSum} task, both the prover and solver functions compute all prime numbers and compute their sum.

The baseline protocol guarantees correctness with absolute confidence by requiring the verifier function $v$ to fully recompute the reference implementation $f$ to validate the solver’s output. However, this approach is highly inefficient: having $v$ duplicate the computation undermines the goal of efficient verification, and nullifies the computational benefits of outsourcing to a solver $s$.

Ideally, a verifiable computing protocol should balance correctness, computational cost, and memory usage by allowing the desired level of confidence to be specified while minimizing the workload of both the solver and verifier. In the following sections, we explore advanced techniques for optimizing verification effort through statistical and cryptographic methods.

\subsection{The SPEX Protocol}
\label{sec:spex}

We introduce \sym{Statistical Proof of Execution (SPEX)}, a sampling-based protocol for verifiable computation.
We outline the requirements for the implementation of the solver $s$ and verifier $v$ functions, including their input and output types, and discuss their implementation for the \sym{PrimeSum} example from Section~\ref{sec:drivingex}.

The solver function $s$ accepts a \textit{SolverRequest} object as input and produces a \textit{SolverResponse} object as output. Similarly, the verifier function $v$ takes a \textit{VerifierRequest} object as input and returns a \textit{VerifierResponse} object as output:

\begin{itemize}
  \item $s(inputs: \textit{ SolverRequest}) \rightarrow \textit{ SolverResponse}$
  \item $v(inputs: \textit{VerifierRequest}) \rightarrow \textit{VerifierResponse}$
\end{itemize}

\subsubsection{Solver function}

The solver function $s$ returns the outputs of task $t$ alongside a cryptographic proof $p$ that might be used to verify its correctness. Its input and output types are defined as follows:

\begin{itemize}
  \item The \textit{SolverRequest} data class contains two fields: \textit{solverInput} and \textit{falsePositiveRate}. The \textit{solverInput} field is task-specific and remains agnostic to our protocol. The \textit{falsePositiveRate} field is an input parameter that specifies the expected rate of false positives in the generated proof $p$.
  \item The \textit{SolverResponse} data class contains two fields: \textit{solverOutput} and \textit{solverProof}. The \textit{solverOutput} field is task-specific and remains agnostic to our protocol, while the \textit{ solverProof} field is a \textit{SolverProof} object.
  \item The \textit{SolverProof} data class represents the proof $p$ and contains two fields: \textit{bloomFilter} and \textit{countItems}. The \textit{bloomFilter} field is a byte string that represents a serialized Bloom filter, while the \textit{countItems} field is an integer that records the number of items added to the Bloom filter.
\end{itemize}
An example of implementation of solver function $s$ for the \sym{PrimeSum} task is illustrated in Algorithm~\ref{alg:primesum-solver}.

\begin{algorithm}
  \caption{\sym{PrimeSum-solver}(request: SolverRequest).}
  \label{alg:primesum-solver}
  \begin{algorithmic}[1]
    \STATE $B \leftarrow \mbox{Bloom-filter}(\mbox{request.falsePositiveRate})$
    \STATE $I \leftarrow \mbox{request.solverInput}$
    \STATE $R \leftarrow 0$
    \FOR{$i \in I$}
    \STATE $P \leftarrow \mbox{nth-prime}(i)$
    \STATE $R \leftarrow R + P$
    \STATE $B.insert(P)$
    \ENDFOR
    \RETURN \mbox{SolverResponse(solverOutput=R, solverProof=($B$,$|I|$))}
  \end{algorithmic}
\end{algorithm}

The \sym{PrimeSum-solver} algorithm computes the sum of prime numbers corresponding to a given sequence of indices while maintaining a lightweight proof of computation using a Bloom filter. It begins by initializing a Bloom filter $B$ with a specified false-positive rate and setting the result accumulator $R$ to zero. For each index $i$ in the input sequence, the algorithm calculates the $i$-th prime number, $P$, adds $P$ to $R$, and inserts $P$ into $B$.
The algorithm returns a response containing the computed sum $R$ and the solver proof.

\subsubsection{Verifier function}

The verifier function $v$ verifies the outputs of the solver function $s$ by fully or partially reproducing its computational states and validating the corresponding cryptographic proof $p$.
It has the following input and output types:

\begin{itemize}
  \item The \textit{VerifierRequest} data class consists of four fields: \textit{solverRequest}, \textit{solverOutput}, and \textit{solverProof}, as previously defined, along with \textit{verificationRatio}, which represents the confidence level $\delta$ as defined in Problem~\ref{problem:1}. \item The \textit{VerifierResponse} data class includes three fields: \textit{isVerified}, a boolean indicating whether the verification was successful, \textit{countItems}, which denotes the number of computational states sampled during the process, and \textit{evidence}, an optional field that is empty by default but may contain information explaining a failed verification.
\end{itemize}

An example of implementation of the verifier function $v$ for \sym{PrimeSum} is illustrated in Algorithm~\ref{alg:primesum-verifier}.

\begin{algorithm}
  \caption{\sym{PrimeSum-verifier-L}(request: VerifierRequest).}
  \label{alg:primesum-verifier}
  \begin{algorithmic}[1]
    \STATE $B \leftarrow \mbox{ request.solverProof}$
    \STATE $E \leftarrow$ estimate false positive rate of $B$ (see Section~\ref{sec:fpr})
    \IF{$E-\epsilon > \mbox{request.solverRequest.falsePositiveRate}$}
    \RETURN \mbox{(isVerified=False, countItems=$0$, evidence=$E$)}
    \ENDIF
    \STATE $I \leftarrow \mbox{request.solverRequest.solverInput}$
    \IF{$|I| \neq \mbox{B.countItems}$}
    \RETURN \mbox{(isVerified=False, countItems=$0$, evidence=$I$)}
    \ENDIF
    \STATE $\delta \leftarrow$ \mbox{request.verificationRatio}
    \STATE $N \leftarrow 0$
    \STATE $I^* \leftarrow \lceil  \delta \cdot |I| \rceil$ random samples without repetition from $I$
    \FOR{$i \in I^*$}
    \STATE $P \leftarrow \mbox{nth-prime}(i)$
    \STATE $N \leftarrow N + 1$
    \IF{\mbox{not} $B.lookup(P)$}
    \RETURN \mbox{(isVerified=False, countItems=$N$, evidence=$P$)}
    \ENDIF
    \ENDFOR
    \RETURN \mbox{(isVerified=True, countItems=$N$)}
  \end{algorithmic}
\end{algorithm}

The \sym{PrimeSum-verifier-L} algorithm initiates the verification of a solver's proof by estimating its false positive rate (FPR) as defined in Section~\ref{sec:fpr}. If the estimated FPR is above the allowed threshold or the expected number of items does not match, the verification process fails. Otherwise, a subset $I^*$ of inputs is sampled, and each item's $n$-th prime is checked in the proof $B$. If any lookup fails, verification stops with a negative result. If all checks pass, the algorithm returns success along with the size of the sampled set.
This verifier is robust against \sym{lazy} solvers, as the primary computational effort lies in generating primes, while their summation is negligible. The verifier mitigates lazy behavior by randomly checking for the presence of the correct primes. However, it cannot prevent an \sym{adversarial} solver from computing all primes, constructing a valid Bloom filter, and still manipulating the reported result.

The primary defense against \sym{adversarial} solvers is to let the verifier independently recompute the output with probability $\delta$ and validate its consistency with the solver's reported result. This verification strategy is implemented in Algorithm~\ref{alg:primesum-verifier2}.

\begin{algorithm}
  \caption{\sym{PrimeSum-verifier-A}(request: VerifierRequest).}
  \label{alg:primesum-verifier2}
  \begin{algorithmic}[1]
    \STATE $\delta \leftarrow$ \mbox{request.verificationRatio}
    \IF{$\textit{Uniform}(0,1) \leq \delta$}
    \STATE $R \leftarrow $ \mbox{PrimeSum-solver(request.solverRequest)}
    \IF{request.solverOutput $\neq$ R.solverOutput}
    \RETURN \mbox{(isVerified=False, countItems=$1$, evidence=$R$)}
    \ENDIF
    \ENDIF
    \RETURN \mbox{(isVerified=True, countItems=$1$)}
  \end{algorithmic}
\end{algorithm}

For a sufficiently large number of solver and verifier task requests, both \sym{PrimeSum-verifier-L} and \sym{PrimeSum-verifier-A} incur a computational overhead proportional to the confidence level $\delta$. \sym{PrimeSum-verifier-L} validates the solver’s proof by randomly sampling inputs and checking their corresponding $n$-th primes in the Bloom filter to detect missing values. Meanwhile, \sym{PrimeSum-verifier-A} strengthens security by probabilistically recomputing the solver’s output and comparing it to the reported result, rejecting mismatches. Increasing $\delta$ raises both the overhead and the reliability of verification.

While both methods operate with confidence level $\delta$ and ensure correctness, \sym{PrimeSum-verifier-L} applies guarantees to each individual solver request, whereas \sym{PrimeSum-verifier-A} enforces them collectively across all requests for the task.

\subsection{Sampling of Computational States}
\label{sec:samcomp}

The \sym{PrimeSum} task served as a driving example for introducing solver and verifier functions via input-output interfaces, alongside an implementation designed to defend against both \sym{lazy} and \sym{adversarial} solvers. However, in a general compute pipeline, defining meaningful computational states that encapsulate both intermediate results and final outputs, while remaining efficient for verification, is still a challenge. This section explores this problem.

Computation can be parallel, sequential, or a combination of both. For example, batch AI inference is a parallel process, where multiple independent inputs are processed simultaneously to generate independent outputs. In contrast, preprocessing followed by AI inference is sequential, as inference must wait for preprocessing to complete.

\begin{figure}[tb]
  \begin{center}
    \includegraphics[width=.8\columnwidth]{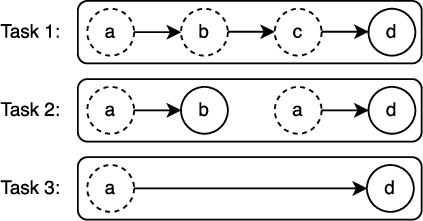}
    \caption{Example of sequential (1, 3) and parallel (2) tasks.}
    \label{fig:task12}
  \end{center}
\end{figure}

As illustrated in Figure~\ref{fig:task12}, Task 1 follows a sequential pattern, while Task 2 operates in parallel. Task 3 represents an extreme case of a sequential task, where no intermediate states are modeled. Here, $a$ serves as the input for all tasks, $d$ is the output of Tasks 1 and 3, and Task 2 produces the output $ (b, d)$. Since all tasks will eventually be re-executed with potentially different inputs, they can all be considered parallelizable across multiple re-executions.

Now, suppose we aim to safeguard a task $t$ against \sym{lazy} solvers while ensuring a confidence level of $\delta = 0.5$.
For Task 1, verification requires multiple task re-executions. A state is randomly sampled for validation—for instance, one verification may check up to state $a$ to ensure correct inputs, while another may verify state $d$, necessitating full reproduction of the execution. Verification is performed by checking whether the sampled state appears in the Bloom filter.
For Task 2, verification is achieved by randomly selecting one of the independent sub-tasks. For example, choosing $a \rightarrow b$, reproducing the computation, and confirming that state $b$ is recorded in the Bloom filter.
In the extreme case of Task 3, where the task consists of a single indivisible step, verification requires a full task re-execution with probability $\delta$, using the Bloom filter to confirm the presence of state $d$. However, in this case, the Bloom filter becomes redundant, as the verifier can directly compare the redetermined output with the original solution.

Similarly, when safeguarding against \sym{adversarial} solvers, validation must be performed without relying on the Bloom filter, instead directly verifying the output, following the a similar verification process used for \sym{lazy} solvers in Task 3.

For each task, the verification effort averages to $50\%$ of the total solver computation time across multiple re-executions.

By carefully designing computational states for verification, the \sym{SPEX} protocol can serve as a generalized verification framework for any computational process and confidence level, ensuring robustness against both \sym{lazy} and \sym{adversarial} solvers, with the computational overhead proportional to the confidence level $\delta$.

\subsection{Bloom Filters as Proof of Execution}

\label{sec:bloom}
Bloom filters are efficient probabilistic data structures designed for fast membership testing. The execution of a function can be represented as reaching multiple computational states, either sequentially or in parallel. By hashing these states, a collection of hash values is generated and stored in a Bloom filter. This filter then acts as a compact, queryable representation of the computation, enabling verification of whether a specific state has been reached without storing the full state.

Unlike Merkle trees, which require path traversal for verification, Bloom filters enable constant-time membership queries and offer greater space efficiency by eliminating the need to track hashes along the traversal path.

The remaining of this section consists of several independent subsections, each addressing a different aspect. The organization is as follows:
Section~\ref{sec:fpr} explores methods to prevent saturated Bloom filters.
Section~\ref{sec:int_fp} explains how to hash integers.
Section~\ref{sec:hashing_fp} explains how to hash arrays of floating-point values while accounting for minor variations.
Section~\ref{sec:semhash} covers semantic hashing techniques for matching semantically similar objects.
Section~\ref{sec:hashagent} discusses approaches for hashing agentic plans.

\subsubsection{False Positive Rate Estimation in Bloom Filters}
\label{sec:fpr}
The false positive rate (FPR) of Bloom filter $B$ is defined as the probability that an element, which was not intentionally inserted into $B$, is mistakenly reported as present. It is estimated
as $\textit{FPR}(B) \approx m / n$, where $m$ represents the number of completely random hashes falsely identified as present in $B$ and $n$ is the total number of random trials.

By configuring and validating the false positive rate, we can prevent \sym{lazy} solvers from flooding the Bloom filter with all possible hashes, which would otherwise result in every lookup yielding a positive result. This safeguards the distinction between valid and invalid computational states, preserving verification integrity.

\subsubsection{Hashing of Integers}
\label{sec:int_fp}

Given an integer $x$, the hashing function \sym{HashInteger}($x$) computes an unsigned 64-bit integer by first representing $(x \mod 2^{64})$ as a little-endian 64-bit unsigned integer, computing the SHA-256 hash of the corresponding byte array, and extracting the first 64 bits of the resulting hash as the output, encoded as a little-endian 64-bit unsigned integer.

\subsubsection{Hashing of Numerical Arrays}
\label{sec:hashing_fp}

In data frames, tables are typically represented as numerical arrays, usually consisting of floating-point numbers, while in machine learning, tensors encode computational states. Hashing arrays of these floating-point numbers is crucial for enabling these applications.

Minor variations in floating-point values can occur due to numerical stability, hardware architecture, GPU non-determinism, or encoding differences. However, applying a small tolerance ensures these values are effectively equivalent.

Algorithm~\ref{alg:hashing-fp} provides an efficient method for encoding numerical arrays into a hashed representation while incorporating a tolerance parameter $\epsilon$.
The process consists of the following steps:

\begin{algorithm}
  \caption{\sym{HashArray}(A, $\epsilon$).}
  \label{alg:hashing-fp}
  \begin{algorithmic}[1]
    \STATE $N \leftarrow$ First significant digit position of $\epsilon$
    \STATE $B = [A - \epsilon,  A + \epsilon]$
    \STATE $Q \leftarrow \lfloor B \times 10^N \rfloor$
    \STATE $H \leftarrow [R_i, ...]$, where $R_i = \{ \sym{HashInteger}(r) : r \geq Q_i^0, r \leq Q_i^1\}$
    \RETURN $H$
  \end{algorithmic}
\end{algorithm}

The algorithm first determines the precision level based on the tolerance parameter $\epsilon$ and defines bounds around each value in the input array $A$, forming the intervals $B$. These bounded values are then quantized by scaling them to the required precision level, ensuring that at least the first significant digit of $\epsilon$ is preserved. The floor function is then applied, storing the quantized values in $Q$. Next, these values are mapped into integer ranges and hashed, producing the list of hash sets $H$. For each value $A[i]$ in the input array $A$, the process generates a corresponding set of hashes $H[i]$. If two arrays $A$ and $A'$ are element-wise equal within the tolerance $\epsilon$, they will always share at least one matching hash in each corresponding set.

A smaller $\epsilon$ increases sensitivity to minor variations in the data, leading to a more precise representation but also a higher risk of false mismatches due to insignificant numerical differences. Conversely, a larger $\epsilon$ allows for greater tolerance to noise or rounding errors but may reduce the ability to distinguish between genuinely different values. The choice of $\epsilon$ should be guided by the specific application. A practical approach is to analyze the distribution of values in $A$ and select $\epsilon$ such that it captures meaningful differences while ignoring insignificant fluctuations.

Hash sets of numerical arrays can be employed in the \sym{SPEX} protocol, where they are generated by the solver $s$ and checked by the verifier $v$.

\subsubsection{Semantic Hashing of Embeddings}
\label{sec:semhash}
\balance 
Quantizing floating-point numbers that represent semantically similar but non-identical objects does not always produce identical hashes. For example, embeddings of text generated by large language models (LLMs) from the same prompt may be close in distance but still result in different hashes.

To address this, we apply semantic hashing to ensure that similar objects share a partially overlapping set of hashes rather than a single identical hash. This is achieved using Locality-Sensitive Hashing (LSH) \cite{jafari2021survey} for metric spaces in Algorithm~\ref{alg:spex-lsh-m}, which maps similar objects to overlapping hash sets,
allowing for a degree of similarity rather than an exact match.

\begin{algorithm}
  \caption{\sym{HashEmbedding-M}(A, V).}
  \label{alg:spex-lsh-m}
  \begin{algorithmic}[1]
    \STATE $D \leftarrow \{ \textit{dist}(A, v) : v \in V\}$ 
    \STATE $I \leftarrow ((j, k): (j, k) \; \mbox{in ascending order}, \; j < k, \; d_j, d_k \in D)$
    \STATE $H \leftarrow \{ \sym{HashInteger}(2 \cdot \textit{idx}(j,k) + \textit{ind}(d_j < d_k)) : (j, k) \in I\}$
    \RETURN $H$
  \end{algorithmic}
\end{algorithm}

The \sym{HashEmbedding-M} algorithm takes two inputs: a query point $A$ (the embedding to be hashed) and a set of vantage points $V$ (a collection of reference embeddings).

First, it computes the set $D$ of distances between $A$ and each $v \in V$ using the \textit{dist} metric function. Next, it constructs an ordered set $I$ of index pairs $(j, k)$, representing the ranking of all distance pairs in $D$, while excluding duplicates and diagonal elements.

Using the index function \textit{idx}, which returns the position of the pair $(j, k)$ in $I$, and the indicator function \textit{ind}, which returns $1$ if the inequality holds and $0$ otherwise, the set of hashes $H$ encodes the relative ordering of distances in $D$. The total number of hash values in $H$ is $(|V|-1)\cdot|V|/2$, corresponding to the number of unique distance pairs between $A$ and the vantage points.

If two query points $X$ and $Y$ represent semantically similar concepts, their distance rankings with respect to the vantage points will be similar, resulting in a high number of overlapping hashes. Given a set of vantage points $V$, the semantic hashes for $X$ and $Y$ can be determined as $H_X = \sym{HashEmbedding-M}(X, V)$ and $H_Y = \sym{HashEmbedding-M}(Y, V)$.
The semantic similarity between $X$ and $Y$ can be measured as the Jaccard index on their respective semantic hashes $H_X$ and $H_Y$:
\begin{equation}
  \label{eq:ssim}
  J(X,Y) =  |H_X \cap H_Y| \;/\; |H_X \cup H_Y|
\end{equation}

Random hyperplanes can be used instead of vantage points in
Algorithm~\ref{alg:spex-lsh-m}. We refer to this variant as \sym{HashEmbedding-H}.
Random hyperplanes partition space using a large number of hyperplanes and metric spaces partition space using the distance to a sufficiently large number of vantage points.
Depending on the context, both methods can be effective options.

Similar to the hashing of numerical arrays, semantic hashing can be applied in the \sym{SPEX} protocol, where a hash set represents a single embedding, generated by the solver $s$ and checked by the verifier $v$.

\subsubsection{Hashing of LLM-Generated Agentic Plans}
\label{sec:hashagent}

An LLM-Generated Agentic Plan is a structured, goal-directed strategy formulated by a Large Language Model (LLM). It consists of logically sequenced steps to achieve a specified outcome, demonstrating agency through purposeful actions.
For example, given the prompt: \textit{"Transfer one BTC to this address if these conditions are met,"} the LLM outputs:
\vspace*{.1cm}
\begin{small}
  \begin{verbatim}
    1. verify_conditions(wallet, conditions)
    2. execute_transaction(signed_tx)
    3. log_transaction(tx_details)    
\end{verbatim}
\end{small}

These steps can be systematically parsed, mapped to structured function calls with corresponding inputs, and hashed using either numerical arrays (see Section~\ref{sec:hashing_fp}) or embeddings (see Section~\ref{sec:semhash}). The resulting hashes serve as a compact representation of the planned execution flow, enabling their use in \sym{SPEX} for computational state verification.

Although reasoning explanations may differ across LLM calls, the underlying programmatic instructions remain consistent, ensuring both determinism and verifiability. This structured approach allows LLM outputs to be reliably transformed into executable and auditable plans.

\section{Conclusion}
\label{sec:conclusion}

As ML/AI-driven decision-making expands, ensuring verifiability is key to trust and accountability. This paper formalizes verifiable computing and introduces SPEX, a sampling-based protocol that is orders of magnitude cheaper, faster, and simpler than existing methods.
We also addressed the challenge of non-determinism in computational states, providing practical strategies for managing common scenarios. By offering a robust framework for verifiable computing, this work advances the development of transparent and reliable systems, facilitating the broader adoption of AI-driven decision-making in a secure and auditable way.

\bibliographystyle{ACM-Reference-Format}
\bibliography{report}

\end{document}